# Pressure Predictions of Turbine Blades with Deep Learning

Cheng'an BAI AND Chao ZHOU*

Turbomachinery Laboratory

College of Engineering, Peking University

Beijing, 100871, China

**Abstract**

Deep learning has been used in many areas, such as feature detections in images and the game of go. This paper presents a study that attempts to use the deep learning method to predict turbomachinery performance. Three different deep neural networks are built and trained to predict the pressure distributions of turbine airfoils. The performance of a library of turbine airfoils were firstly predicted using methods based on Euler equations, which were then used to train and validate the deep learning neural networks. The results show that network with four layers of convolutional neural network and two layers of fully connected neural network provides the best predictions. For the best neural network architecture, the pressure prediction on more than 99% locations are better than 3% and 90% locations are better than 1%.

**Nomenclature and Abbreviations:**

$c$: Chord

$C_p$: Static pressure coefficient $(p - p_1)/\frac{1}{2}\rho U^2$

$C_x$: Axial position coefficient $x/c$

$d$: Depth of convolution kernel

$p$: Static pressure

*U*: Inlet velocity

*x*: Axial position

**Greek Symbol:**

*ρ*: Density

**Subscript:**

1: Inlet

## INTRODUCTION

The concept of neural network was first brought in 1940s based on the description of neurons in neurophysiology[1]. In 1960s, studies on visual cortex revealed that visual processing has multilayers related to the neural network [2]. In 1980s, Fukushima built the first convolutional neural network (CNN) [3], which was able to perform basic image recognition.

In recent years, convolutional neural networks, one of the artificial neural networks, have been developed and have been proven very effective in many areas such as image classification and game of go. In 2012, CNN was firstly used in the "ImageNet LSVRC contest" and achieved a significant improvement in prediction, which reduced the error rate from 26% in the year 2011 to 15% [4]. The architecture of CNN is further developed and its error in image recognition is further reduced to 2.3% in the year of 2017 [5], which is lower than that of the human beings [6]. Google's AlphaGo, which was also developed based on deep convolutional neural network, beat go master Lee Se-dol in 2016[7]. These amazing achievements drive more and more studies on the application of CNN in many other fields. In the area of aerospace, deep learning methods are being studied in flight control[8], airfoils design for aircrafts[9] and anomaly detection[10].

There is no publication about the deep learning method in the field of turbomachinery so far. This paper will present a study by using the deep learning methods to predict the pressure distribution of two dimensional turbine blades.

**Library of Turbine Airfoil**

Deep learning methods require big data to train the neural networks. The profiles of the library of turbine airfoils are built based on a datum 2D turbine blade, as shown in Figure 1. The main parameters of the datum blade is listed in Table 1. By changing the distance from the blade surface to the chamber line, the turbine blade library was obtained. The main changes occur on the pressure side and the suction side of the blades, and the changes in the geometry near the blade leading edge and the trailing edge are small. All of the blades have almost identical inlet and exit flow angle.

A blade-to-blade method, MISES[11], is used to obtain the pressure distribution on each blade. This solver is based on inviscid/viscous interaction, with integral boundary layer equations involved in boundary layer and wake development.

| Design inlet flow angle | 41.08° |
|---|---|
| Blade exit flow angle | 69.25° |
| Pitch/Chord ($s/C$) | 0.79 |

Table 1 Main parameters of the turbine blades

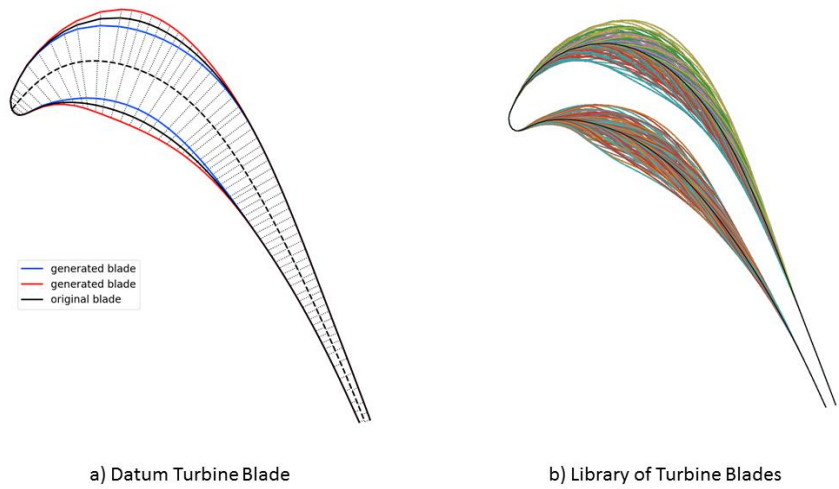

Figure 1: Datum and Library of Turbine Blades

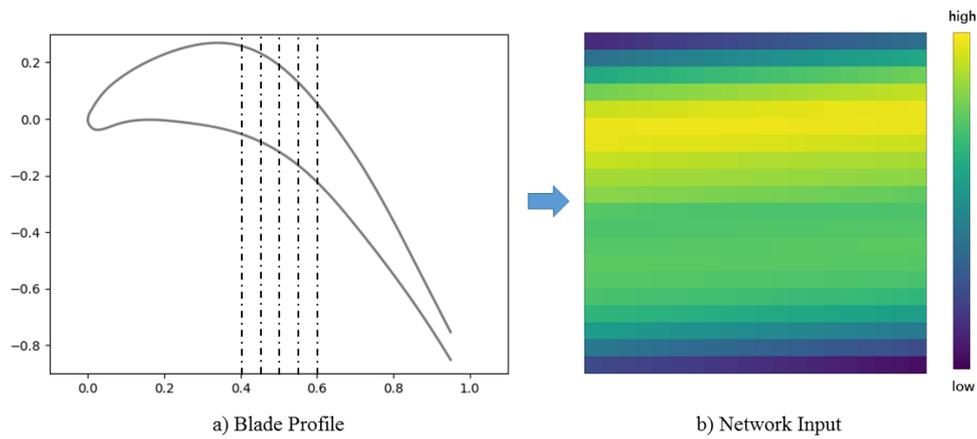

Figure 2: Example of Blade Profile and its Format for Deep Neural Network Input

A library of 63,450 turbine blades and their static pressure distributions are obtained. A total of 400 points are used to define the shape of each 2D turbine blade. For the input of convolutional neural networks, the coordinates of a blade are normalized and arranged to a matrix of 20×20, as shown in Figure 2b. The position of each pixel defines the axial coordinate of the point on the blade, and its value defines the tangential coordinate. In fact, Figure 2b is how the deep neural networks "see" the geometry of a turbine blade.

Figure 3 shows the static pressure distribution on a typical blade surface. The continuous curve is the results of CFD calculation. The value of each point on this curve can be extracted to train the deep neural networks. At each axial position, the values of static pressure for all of the blades in the library are examined and the range are determined. Then, the values are classified according to the possible range of the static pressure coefficient with an interval of 0.1, which will later be used to train the deep neural networks with the classification method. The interval of the label also determines the resolutions of the values for training. After a deep neural network is trained, the output of the static pressure coefficient by the deep neural network is also in the form of labels. For example, at a location of 0.2 Cx on the pressure side, the variation of static pressure coefficient is within the magnitude of 0.5, so it is classified into five labels. On the blade suction surface, the variation of Cp for all the blade profiles in the library is 3.2, so it covers 32 labels.

To predict the Cp value on a certain axial location of the blade profiles, the Cp values on this axial location of a number of blades will be used to train the neural network. The axial coordinate on either the blade pressure side or the suction side is used as a variable. 2/3 of the data in the library are used as the training data, 1/6 of the data are used as the validation data, and the remaining 1/6 data are used as test data.

The prediction of Cp on each location is independent. Considering the requirement of the computational resource, the static pressure distributions on 9 positions on each side of the blade are selected to train deep neural networks. Later, the static pressure distributions on these 18 positions will be predicted by the neural networks. Similar method can be used for more positions.

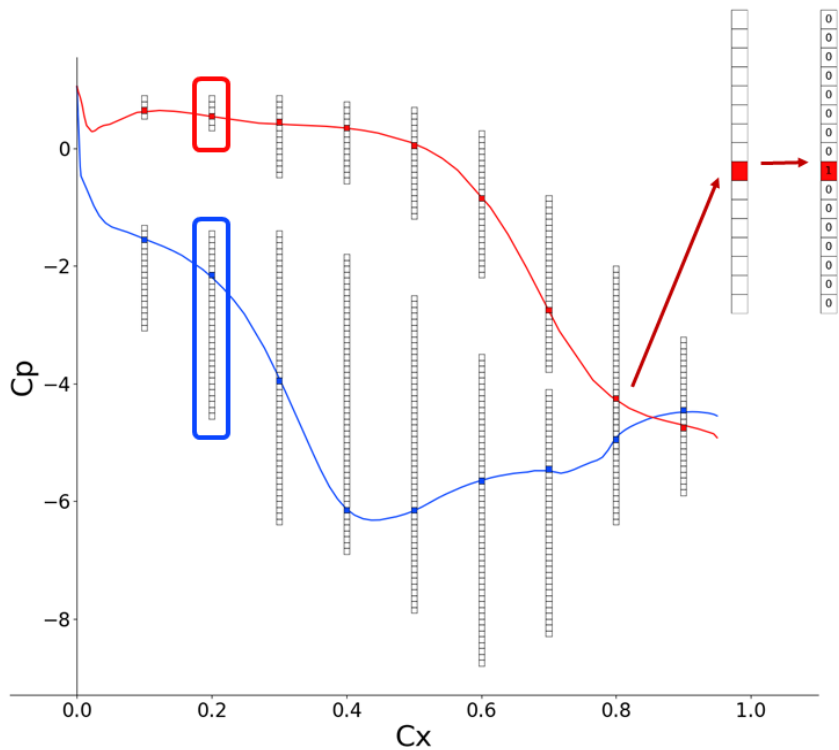

Figure 3: An Example of Cp Distribution and its Classification

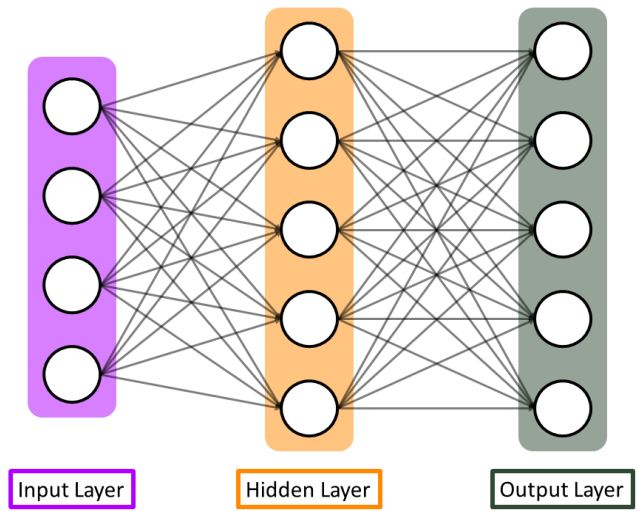

Figure 4: Structure of Two-Layer Fully Connected Neural Network

## Structures of Deep Neural Networks

Deeping learning is a subfield of machine learning consisting of multiple layers of nonlinear data processing[12]. There are two main types of deep learning manners, supervised and unsupervised.

In this paper, supervised learning is used. The deep neural network architecture used in the current study is similar to those used for the recognition of images. The number of convolutional layers and depth of each convolutional layer have crucial effect on the accuracy of classification. The networks with more layers and depth are often able to deal with more complex problems, but they also bring problems such as overfitting and choose of hyperparameters.

The deep neural networks used in the current study are built with TensorFlow[13]. Three architectures of neural network are studied in this paper.

1) 2-layer fully connected neural networks (nn2).

2) 2-layer Convolutional neural networks+ 2-layer fully connected neural networks (cnn2_nn2).

3) 4-layer Convolutional neural networks+ 2-layer fully connected neural networks (cnn4_nn2).

The input of a fully connected neural network (NN) is a vector. The input of a convolutional neural network (CNN) is a 2-D matrix, as shown in Figure 2. The output of CNN and NN are both vectors, and the length of the output vector equal to its number of classifications. The element with the highest value in the output vector is the predicted classification. Here, the Cp is classified, and output is the class that have the highest value.

A full connected neural network (NN) is composed of an input layer, hidden layers, and an output layer. The current NN has one hidden layer, as shown in Figure 4. The output of the hidden layer $h_{W,b}$ is:

$$h_{W,b} = f(W \cdot \vec{x} + \vec{b})$$

Assuming that the input $x$ is a n-dimensional vector, and the output $h_{W,b}$ is a m-dimensional vector, then the weights $W$ is a $m \times n$ matrix and the bias $b$ is a m-dimensional vector. The output size

*m* of each layer is a changeable parameter. *f* is called activation function. Relu[14], Softmax[15] and Dropout[16] are used as activation functions in the fully connected neural networks. The activation function max-pooling[17] is also used.

The key to improve the accuracy of prediction is the use of convolution method to build a convolutional neural network (CNN). Figure 5 demonstrates how the convolution method works. The input is a 2-D matrix called input feature map, shown as the blue part. As illustrated in Figure 5, a kernel slides over the input feature map. At each location, the product of the overlapped part of kernel and input feature map is calculated and it will be the output element of that location[17]. The kernel is iterated and updated.

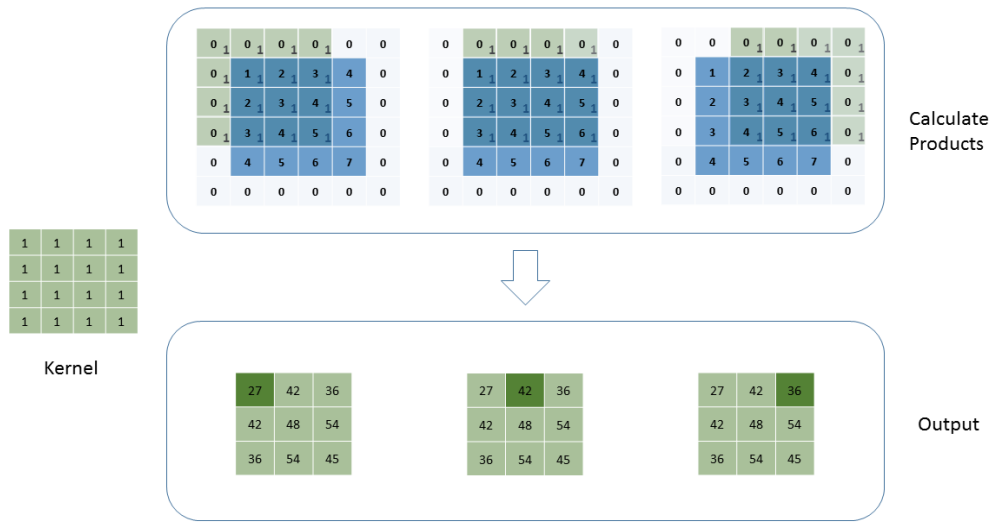

Figure 5: Demonstration of Convolution

The structure of the 2-layer fully connected neural network (nn2) is shown in Figure 6. Activation function Relu and Dropout are used in the first layer. The structure of 2-layer convolutional neural network with 2-layer fully connected neural network (cnn2_nn2) is shown in Figure 7. Activation function Relu and Max Pooling are again used in convolutional layers. The structure of 4-layer convolutional neural network with 2-layer fully connected neural network

(cnn4_nn2) is similar with cnn2_nn2. The difference is that Max Pooling is used every two convolutional layers in cnn4_nn2.

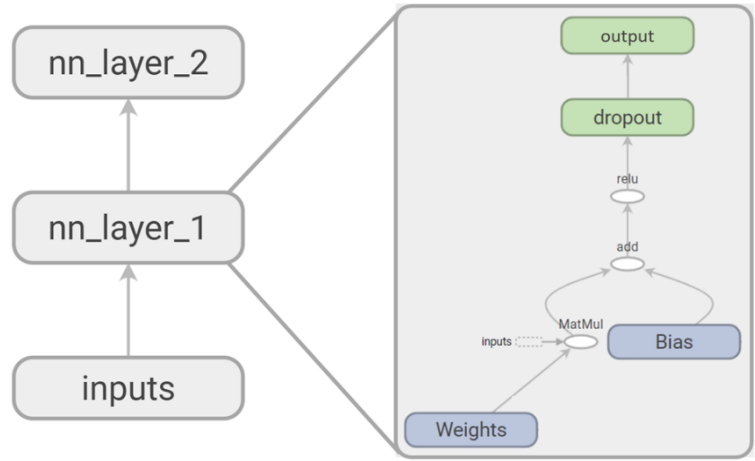

Figure 6: Structure of 2-layer Fully Connected Neural Networks (nn2)

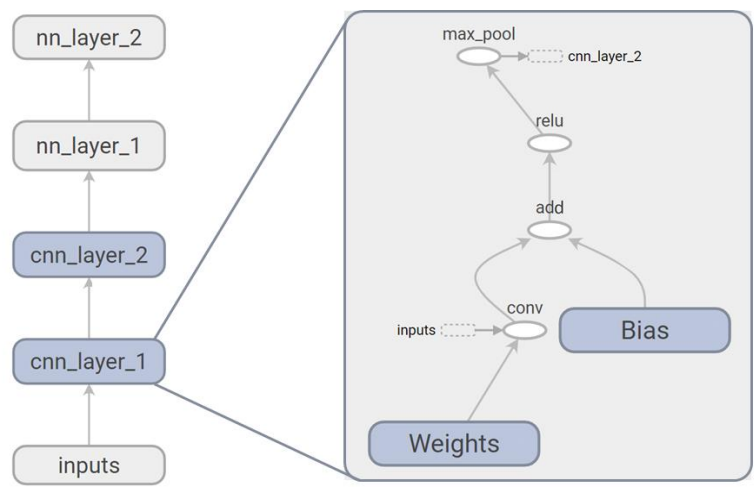

Figure 7: Structure of 2-layer Convolutional Neural Networks+ 2-layer Fully Connected Neural Networks (cnn2_nn2)

## Results and analysis

The method of classification is used in the deep neural networks for the prediction of the pressure coefficient. Each class shown as a label and is enlarged in Figure 8, which covers a range of pressure coefficient equals to about 2% of the exit dynamic head. In other words, the resolution of the static pressure distribution prediction is 2% of the exit dynamic head. This resolution can be improved by reducing the range of pressure distribution covered by each class at a cost of increasing in the use of computational resources.

The prediction of the deep learning networks is compared with the output of the CFD prediction. If the CFD result of the pressure coefficient on one position falls into the class of neural network prediction, the accuracy of the prediction is regarded as 1%. If the CFD result falls into the class next to it, the error increase to 3%, and so on.

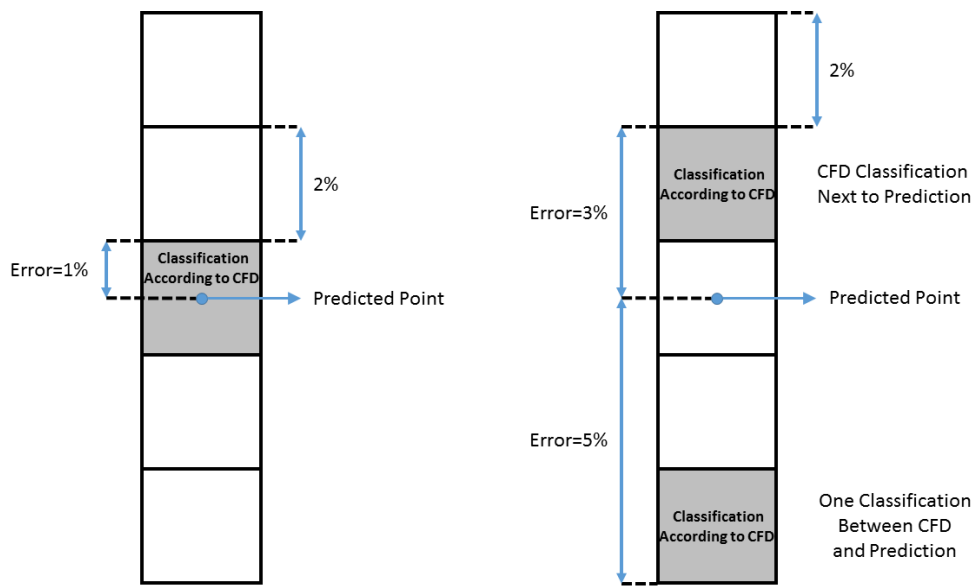

Figure 8: Error Analysis Based on Labels

The accuracy of prediction by deep learning networks at different locations on the blade surface will be examined. Figure 9 shows the percentage of predictions that achieves an accuracy better than 1%. In general, the deep learning networks with CNN architectures achieves much better prediction than the NN networks. For the nn2 network, in which the CNN architectures are not used, the lowest accuracy of prediction appearing on the pressure side is at the location of 0.7Cx, where only about 55% of the cases achieves an accuracy of 1%. On the suction side, the accuracy becomes even worse. For CNN networks, the accuracy of prediction is much higher. The cnn4_nn2 neural network achieves better prediction than the cnn2_nn2 neural network. In general, the CNN networks with more convolutional layers and more kernels achieve better performance. The cnn4_nn2 network with more than 16 kernels achieves very good prediction accuracy. The accuracy of the prediction is low on around 0.4 Cx of the suction surface of the blade, where the flow separation occurs and the variation of the pressure is high. Nevertheless, still more than 90% of cases achieves an accuracy within 1%. The overall prediction is shown in Table 2. For the cnn4_nn2 network, 90% predictions achieves an accuracy better than 1%, and 99% prediction achieves an accuracy better than 3%.

|   | Neural network | No. of Kernels | Accuracy 1% | Accuracy 3% |
|---|---|---|---|---|
| 1 | cnn2_nn2 | 16 | 78% | 97% |
| 2 | cnn2_nn2 | 64 | 86% | 99% |
| 3 | cnn4_nn2 | 16 | 90% | 99% |
| 4 | cnn4_nn2 | 64 | 90% | 99% |

Table 2: Accuracies of Different Networks

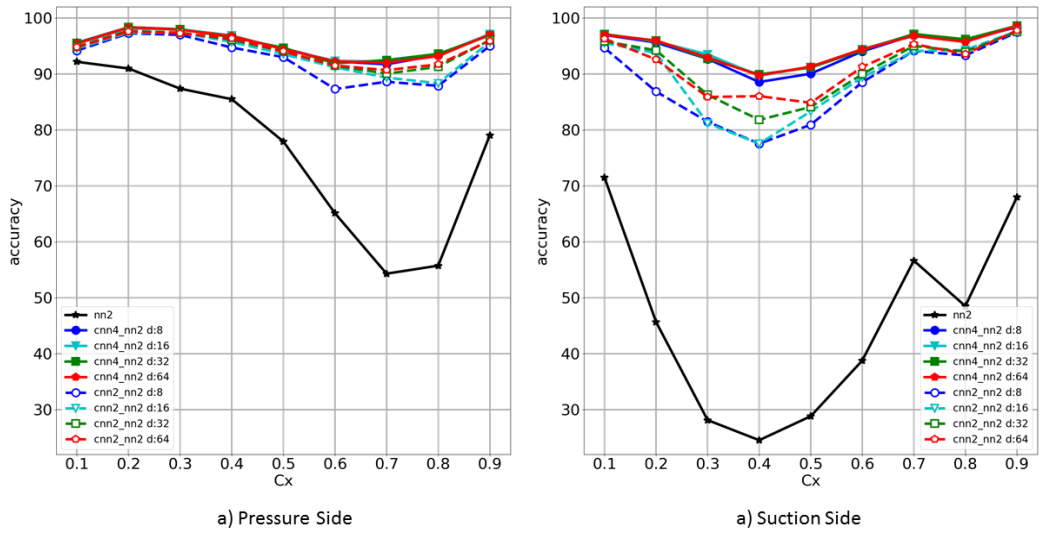

Figure 9: Percentage of Predictions with an Accuracy Better than 1% for Different Deep Networks

An example of prediction is shown in Figure 10. Figure 10a shows the blade profile and Figure 10b shows the result predicted by CFD method and the neural network. The label shows that the CFD and neural network provide the same prediction, and the accuracy of prediction on most position is better than 1%. On the suction surface with Cx of 0.3, the predicted label by neural network is next to the CFD, and the accuracy of prediction is 3%.

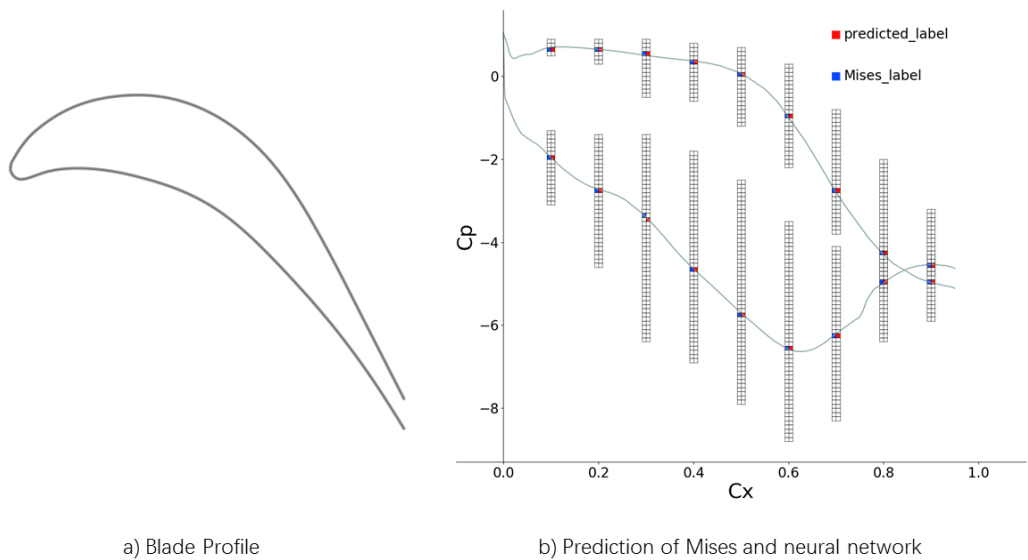

Figure 10: An Example of Prediction

Although deep networks are able to prediction on problems with high accuracy, the mechanism of how it works is not well understood. To analyze the factors that affect the accuracy of the output, the results on the 40% axial chord of the suction surface by the cnn2_nn2 with 16 kernels are presented. A total of 16 convolution kernels with 5*5 are used for the first CNN layer as shown in Figure 7 as 'cnn_layer_1'. The output of the first layer is 16 matrixes with the size of 20*20 as shown in Figure 11b. Traces of horizontal lines can be observed as the format of the input coordinate data.

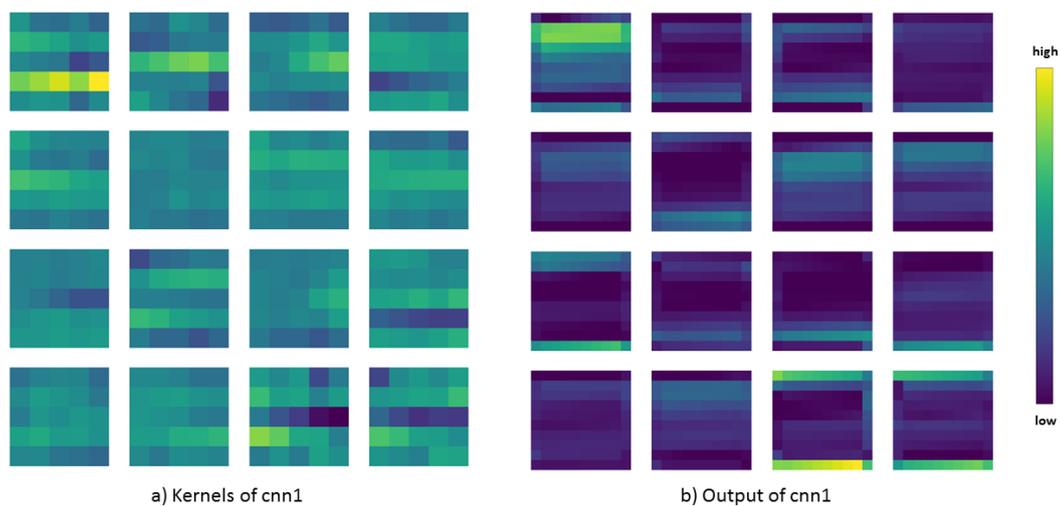

a) Kernels of cnn1        b) Output of cnn1

Figure 11: Example of Kernels and Output of the First CNN Layer

For a cnn2_nn2, another convolution calculation is carried out based on the output of the first CNN layer. The kernels were obtained after several iterations. One example of the kernels is shown in Figure 12a, where only certain part of this kernel have values. In the output of the second convolution shown in Figure 12b, the values in certain areas are higher than the rest of the output, and these area are the key for the activation of the neural network. The following fully connected neural network will provide the activated points based on these activated points.

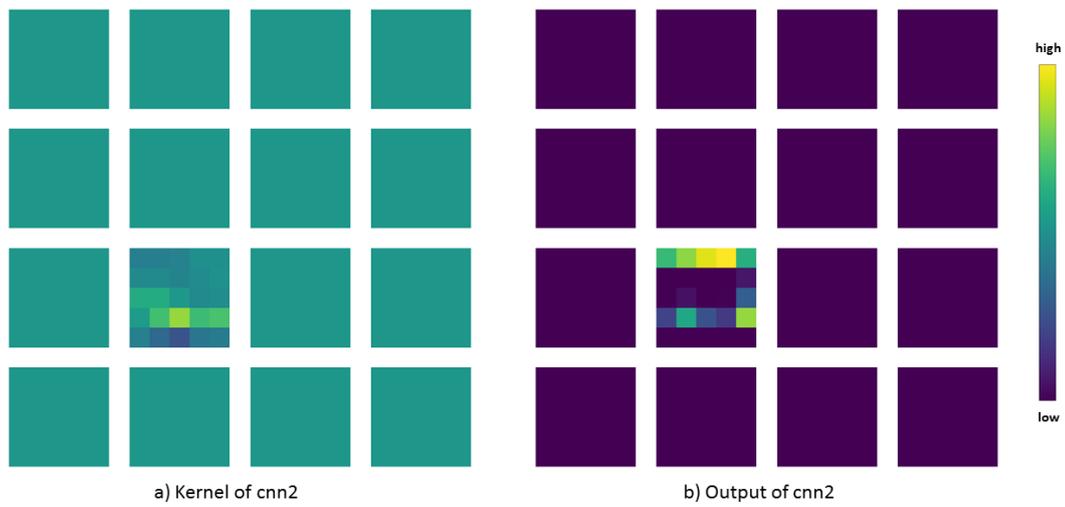

Figure 12: Example of Kernels and Output of the Second CNN Layer

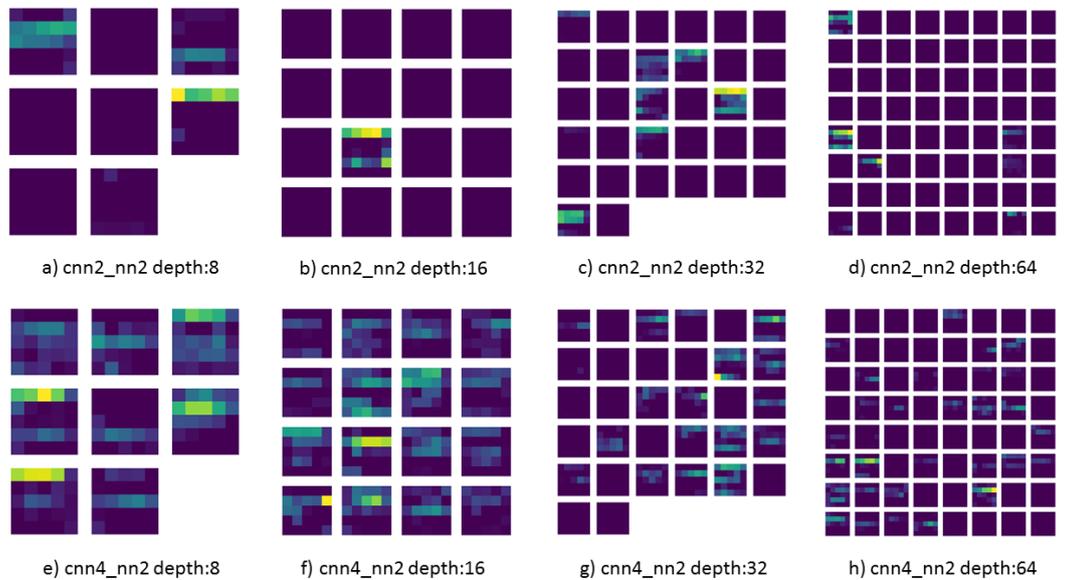

Figure 13: Example Outputs of the Last CNN Layer of Different Networks

Figure 13 shows the examples of outputs of the last convolutional layers in different neural networks. The accuracy of prediction is related to the architecture of the neural networks and the number of activated cells. The cnn2_nn2 depth:8 and cnn2_nn2 depth:16 provide results with an accuracy about 82%, which is low. In Figure 13a and b, the numbers of activated cells are between 10 and 20. Increasing the depth of cnn2_nn2 increases the numbers of activated cells and the accuracy improves. However, the activated cells for cnn2_nn2 depth:64 (Figure 13d) is less than

cnn2_nn2 depth: 32, but the prediction accuracy of cnn2_nn2 depth:64 is better. This is probably due to the extraction of information is better with higher depth.

In the cnn4_nn2 neural networks, more cells are activated. Also, as the depth of the neural network is increased, more cell is activated. However, for the cnn4_nn2 network, the improvement in accuracy by increase the depth is marginal.

**Conclusions and Discussion**

The current study uses deep neural networks to prediction the static pressure distribution of a turbine blade. A library of static pressure distributions of turbine blades are obtained and used to train and validate the deep neural networks. Three different deep neural networks, namely nn2, cnn2_nn2 and cnn4_nn2 are built. The results show that with the convolutional method, the accuracy of deep neural networks in predicting the blade surface static pressure distribution is increased. The cnn4_nn2 neural network provides the best prediction, the pressure prediction on more than 99% locations are better than 3% and 90% locations are better than 1%. It is observed that the accuracy of prediction is related to the number of activated cells, which is affected by the depth and number of convolutional layer.

This study shows that deep neural networks are able to provide fair accuracy in the prediction of static pressure of blade. According to the authors' knowledge, this is the first attempt of using deep neural network in the field of turbomachinery. There are many possible ways to improve the capability deep neural networks for turbomachinery application. We hope the current publication can inspire more works in this area.